\documentclass{article}

\PassOptionsToPackage{sort&compress,square,numbers}{natbib}
     \usepackage[preprint]{neurips_data_2023}
\usepackage[utf8]{inputenc} % allow utf-8 input
\usepackage[T1]{fontenc}    % use 8-bit T1 fonts
\usepackage{hyperref}       % hyperlinks
\hypersetup{
    colorlinks,
    citecolor=black,
    filecolor=black,
    linkcolor=black,
    urlcolor=teal,
}                           % remove boxes / color around links
\usepackage{url}            % simple URL typesetting
\usepackage{booktabs}       % professional-quality tables
\usepackage{amsfonts}       % blackboard math symbols
\usepackage{nicefrac}       % compact symbols for 1/2, etc.
\usepackage{microtype}      % microtypography
\usepackage{xcolor}         % colors
\usepackage{graphicx}
\usepackage{cleveref}       % smart cross-referencing

\title{Into the Single Cell Multiverse:\\an End-to-End Dataset for Procedural Knowledge Extraction in Biomedical Texts}

\author{%
  Ruth Dannenfelser$^{1}$\\
  \And
  Jeffrey Zhong$^{1}$\\
  \And
  Ran Zhang$^{2}$\\
  \And
  Vicky Yao$^{1}$\thanks{Address correspondence to: vy@rice.edu}\\
  \AND
  \vspace{-2.5em}
  \\
  $^{1}$   Department of Computer Science, Rice University\\
  $^{2}$   Department of Genome Sciences, University of Washington
}

\begin{document}

\maketitle
\setcounter{footnote}{0}

\begin{abstract}
  Many of the most commonly explored natural language processing (NLP) information extraction tasks can be thought of as evaluations of declarative knowledge, or fact-based information extraction. Procedural knowledge extraction, i.e., breaking down a described process into a series of steps, has received much less attention, perhaps in part due to the lack of structured datasets that capture the knowledge extraction process from end-to-end. To address this unmet need, we present FlaMB\'e (Flow annotations for Multiverse Biological entities), a collection of expert-curated datasets across a series of complementary tasks that capture procedural knowledge in biomedical texts. This dataset is inspired by the observation that one ubiquitous source of procedural knowledge that is described as unstructured text is within academic papers describing their methodology. The workflows annotated in FlaMB\'e are from texts in the burgeoning field of single cell research, a research area that has become notorious for the number of software tools and complexity of workflows used. Additionally, FlaMB\'e provides, to our knowledge, the largest manually curated named entity recognition (NER) and disambiguation (NED) datasets for tissue/cell type, a fundamental biological entity that is critical for knowledge extraction in the biomedical research domain. Beyond providing a valuable dataset to enable further development of NLP models for procedural knowledge extraction, automating the process of workflow mining also has important implications for advancing reproducibility in biomedical research.
\end{abstract}

\section{Introduction}
The recent onslaught of pre-trained language models has spurred on tremendous advances in a range of natural language processing (NLP) applications, including named entity recognition (NER), named entity disambiguation (NED), sentiment analysis, and relation extraction \cite{nadeau_survey_2007,li_survey_2022,cucerzan_large-scale_2007,zhang_deep_2018,zhang_end--end_2017}. These applications mostly fall under the umbrella of tasks that aim to extract \textit{declarative knowledge}, sometimes also referred to as ``knowing that,'' since these tasks focus on matters of factual knowledge (e.g., \textit{knowing that} ``neuron'' is a cell type) \cite{ryle_knowing_1945,wisdom_concept_1949}. Declarative knowledge is often contrasted with \textit{procedural knowledge}, or ``knowing how,'' (e.g., \textit{knowing how} to conduct an experiment) \cite{ryle_knowing_1945,wisdom_concept_1949}. Early AI researchers raised the importance of developing representations of procedural knowledge, given that performing plans or procedures is a fundamental way in which humans navigate the world \cite{georgeff_procedural_1986}. However, compared with declarative knowledge extraction, there remains a vast gap in the development and application of machine learning methods towards procedural knowledge tasks \cite{mujtaba_recent_2019}.

Recently, there has begun to be a renewed interest in using machine learning to model procedural knowledge, especially knowledge extraction from text using NLP. These efforts have mostly focused on cooking and other common household tasks \cite{min_survey_2019,chu_distilling_2017}, business processes \cite{bellan_pet_2023}, and technical manuals or manufacturing \cite{gupta_mining_2018}. The specific applications that have garnered interest seem to have been naturally motivated by either the emergence of valuable datasets (e.g., online recipes for cooking, WikiHow for various how-to tasks) or economic gain through business process optimization. Interestingly, one of the main ways scientists and engineers communicate their findings---through academic papers---is a prime source of unstructured text describing ``know-how,'' yet few studies explore extracting procedural knowledge from scientific literature. This is the case though there is also an abundance of open access scientific literature that is frequently used for many standard declarative knowledge extraction studies.

We posit that there are 3 main reasons that procedural knowledge extraction from scientific literature is not currently widely studied:
\begin{enumerate}
    \item Though most research papers will describe procedures, i.e., methods, they are typically not written with as much structure as a recipe or technical manual, and thus not as easy to model ``off the shelf.'' In fact, methods sections are often organized by thematic categories and do not necessarily represent the ``temporal ordering'' in which the individual steps were done.\footnote{Note that here, temporal ordering is used loosely, as we are simply referring to the workflow ordering that a reader can deduce from the manuscript. It is of course common that scientific manuscripts present their main results in differing order than originally conducted. That said, we expect that internal ordering of tasks within each major result to be typically a good reflection of what was actually performed.} It is also often the case that the results sections need to be read together with the methods sections to reconstitute how various tools were used. 
    \item There can be varying degrees of ambiguity in a scientific manuscript when systematically describing a workflow. The same method or software tool can be used at several time points throughout a paper, but in different contexts and for different purposes. For example, principal component analysis (PCA) can be used for dimensionality reduction, feature selection, or visualization. Failure to account for context may lead to a workflow that appears to simply have a chain of PCAs. In addition, multiple parallel workflows can be described in a single paper. For example, a single paper can consider multiple datasets, each of which are processed differently, before they are analyzed jointly.
    \item Unlike writing down recipes or household tasks, annotating the workflow used in a scientific paper is challenging without domain expertise, thus resulting in a bottleneck for developing structured datasets.
\end{enumerate}

Motivated by these observations, we introduce FlaMB\'e (Flow annotations for Multiverse Biological entities). FlaMB\'e is a collection of structured annotations in biomedical research papers, with a particular focus on computational analysis pipelines in single cell research. While scientists have long been interested in studying single cells \cite{nature_methods_method_2014,eberwine_promise_2014}, it was with the introduction of high-throughput single cell sequencing technologies around 2010 \cite{tang_mrna-seq_2009} that this area has exploded in activity, not only in applications of this experimental technique to various biomedical applications, but also in the development of computational tools and software to analyze the resulting data. Recent efforts to wrangle the space of analysis tools has resulted in specialized databases such as scRNA-tools \cite{zappia_exploring_2018}, which currently tracks over 1,500 software tools across over 30 analysis tasks.\footnote{The terminology scRNA-tools uses for these analysis tasks is ``categories,'' since they are focused on grouping tools by their applications. We simplify the terminology here to make clear that each tool can have multiple category tags.} Interestingly, the majority of tools catalogued by scRNA-tools are used for more than one analysis task, and one of the most commonly used tools, Seurat \cite{satija_spatial_2015}, is associated with as many as 10 categories of tasks, further highlighting the importance of considering context. 

\begin{figure}[htbp]
  \centering
  \includegraphics[width=9.5cm]{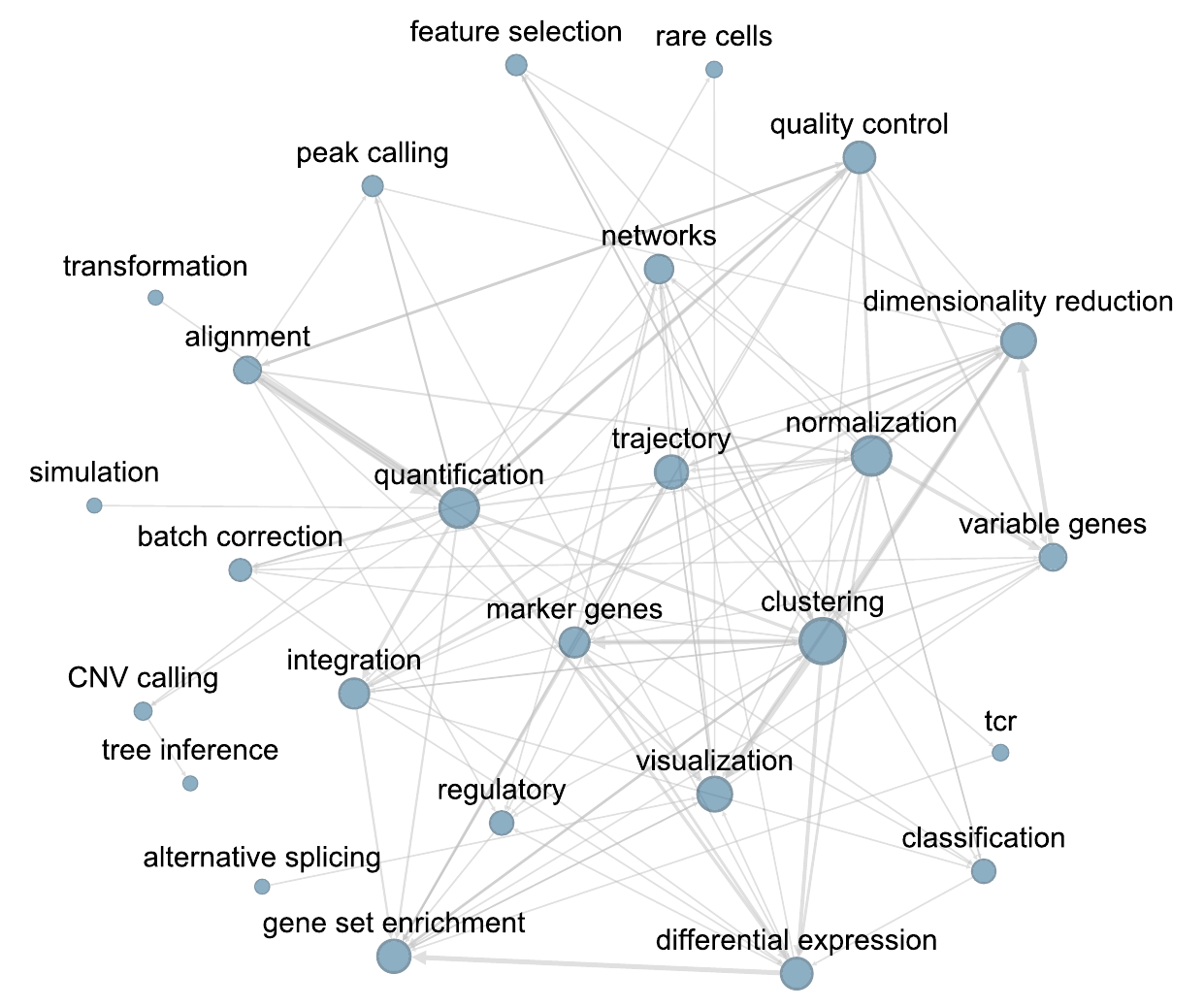}
  \caption{\textbf{Example overview of the workflow of tool contexts (analysis tasks).} Summary figure of workflows from different tool contexts captured by FlaMB\'e. Direction of edge represents which analysis task was completed prior to the output being sent to the following task. Weight of edges represent the number of papers that mentioned. Node size corresponds to degree, i.e., the number of papers that mentioned the corresponding analysis task.}
\vspace{-2em}
\end{figure}

In FlaMB\'e, we develop a structured representation of the procedural knowledge represented in scientific literature by considering (1) the \textit{targets} of the study, which in the case of single cell research, are the tissues and/or cell types that are assayed; (2) the \textit{tools} applied in the study as well as the analysis task or \textit{context} in which they are being used; and (3) the \textit{workflow} between tools and analysis tasks, e.g., when PCA is used for dimensionality reduction before the results are clustered using DBSCAN. Part of the motivation in structuring FlaMB\'e in this manner is that we can break down the more complex, unstructured goal of procedural knowledge extraction into existing, more manageable declarative knowledge extraction tasks. For example, the identification of targets and tools in text reduces to NER and NED tasks. 

Overall, we present 55 full text papers, including over 420,000 tokens, annotated for relevant entities and relations from the PubMed Central Open Access Subset by domain experts (computational biologists). To improve coverage over a more diverse set of journals and entities, we also provide tissue/cell type annotations in 1,195 paper abstracts mined from PubMed, covering over 270,000 tokens. The entire dataset provides entity annotations as well as disambiguation, where entities are linked to identifiers in relevant knowledge bases. To our knowledge, FlaMB\'e is the largest NER and NED dataset for tissues/cell types. Furthermore, we also provide annotations for software tools and computational methods, also capturing 28 unique contexts in which the tools are used for single cell research and nearly 400 workflow relations between (tool, context) pairs. An example visualization of the flow between contexts is shown in Fig. 1. FlaMB\'e is available for exploration and download at \url{https://github.com/ylaboratory/flambe}.

We illustrate some example use cases for FlaMB\'e here, but the richness of this dataset has many more potential downstream applications in machine learning as well as computational biology and the wider biomedical field. In general, the complexity of working with single cell data and its capacity for a variety of different workflows, together with its important biomedical applications, is ultimately what led us to choose the area of single cell research for FlaMB\'e. However, we have also proposed a systematic framework to distill procedural knowledge into a structured dataset in a manner that considers some of the unique challenges of scientific literature. It is our hope that FlaMB\'e provides a useful foundation for future ``science-know-how'' modeling and datasets.

\section{Related Work}
FlaMB\'e is designed to represent a collection of complementary tasks that together form the basis of a structured representation that captures procedural knowledge in biomedical texts. Here, we discuss related datasets and research efforts.

\paragraph{Biomedical NLP}
Systematic evaluations of language models in a variety of different benchmarking efforts have revealed that for specialized domains like biomedicine, language models developed using domain-specific text (e.g., scientific literature) often outperform general-domain language models (e.g., trained on Wikipedia, news articles, webpages, etc.) on domain-specific tasks \cite{beltagy_scibert_2019,gu_domain-specific_2021,lee_biobert_2020,peng_transfer_2019}. Furthermore, it seems that mixed-domain pretraining can sometimes hurt more than help, suggesting that transfer learning is at times unsuccessful due to how different general-domain text is from biomedical text \cite{gu_domain-specific_2021}. In general, there has been a demonstrated need for both domain-specific pretrained language models as well as domain-specific datasets for method benchmarking. 

In the biomedical domain, language models are often trained on a mix of abstracts from PubMed and full text articles from PubMed Central \cite{lee_biobert_2020,kanakarajan_bioelectrapretrained_2021,beltagy_scibert_2019,gu_domain-specific_2021}, at times also with additional scientific text such as medical records \cite{peng_empirical_2020}. A variety of biomedical NLP benchmarking datasets have also been developed \cite{peng_empirical_2020,gu_domain-specific_2021,weber_hunflair_2021}, but often individual tasks can be fragmented. Very recently, large scale efforts like BigBio \cite{fries_bigbio_2022} have systematically organized comprehensive public collections of biomedical NLP datasets. BigBio's curation revealed that the largest represented task within biomedical NLP is unsurprisingly NER, as there are a variety of biological entities that are often of interest for text mining (e.g., diseases, gene names, chemical compounds, anatomy/tissue/cell type). Of particular relevance to our work here are previous dataset curation efforts for tissue/cell type \cite{ohta_open-domain_2012,pyysalo_anatomical_2014,pyysalo_overview_2013,neves_annotating_2012}. However, not only are these datasets smaller in terms of total annotations in comparison with FlaMB\'e, but furthermore, none provide NED. Disambiguating these terms and linking them to a systematic knowledge base provides more utility for the biomedical community and also enables incorporation of information from the associated knowledge base for improved knowledge extraction.

The other entity that has recently begun to be considered for NER in biomedical literature is software. As the field of computational biology grows and, accordingly, the number of software tools and computational methods, systematic identification and analysis of tool usage has become more relevant. Large-scale curation efforts for NER and NED here include bioNerDS \cite{duck_bionerds_2013}, SoftCite \cite{du_softcite_2021}, and SoMeSci \cite{schindler_somesci-_2021}. These previous datasets have differing limitations. Both bioNerDS and SoftCite only consider articles published before 2011\footnote{bioNerDS further restricts its annotations to only two journals, \textit{BMC Bioinformatics} and \textit{Genome Biology}.} in their dataset, while SoMeSci curates articles as recent as 2020. However, SoMeSci's main endpoint is a knowledge graph and thus does not provide its annotations in an easily usable format. Both SoftCite and SoMeSci have been used as training data to automatically identify software mentions across millions of scientific articles \cite{istrate_large_2022,schindler_role_2022}, though the resulting automatically annotated datasets differ greatly. Finally, all previous datasets focus solely on software. Because one of the key goals of FlaMB\'e is to extract data processing and analysis workflows, we also wanted to expand annotations to computational methods that are often referred to in scientific papers without necessarily a specific associated software (e.g., PCA, SVM).

\paragraph{Procedural knowledge extraction}
Recent efforts in procedural knowledge extraction have been spurred on by the increasing availability of naturally arising procedural knowledge-related data sources. In fact, the widespread availability of online recipes have given rise to the new research area of ``food computing.'' \cite{min_survey_2019} Other areas where there is active research in procedural knowledge extraction include household tasks based on mining data sources such as WikiHow, Instructables, and eHow \cite{chu_distilling_2017}, technical manuals \cite{gupta_mining_2018}, and business processes \cite{bellan_pet_2023}.

There has also been some limited attempts to examine scientific literature as an application area. \citeauthor{song_procedural_2011} propose representing procedural knowledge as (target, action, method) triplets based on MEDLINE abstracts \cite{song_procedural_2011}, and \citeauthor{halioui_bioinformatic_2018} consider using process-oriented case-based reasoning to extract workflows from papers mentioning phylogenetic analyses from PubMed \cite{halioui_bioinformatic_2018}. Interestingly, these two pieces of work fall on two ends of the spectrum in terms of the complexity of the representations they propose. In addition to the limitations of modeling an entire workflow from only an abstract, \citeauthor{song_procedural_2011}'s proposed representation is also unable to take into account when tools are applied in different contexts. Meanwhile, \citeauthor{halioui_bioinformatic_2018}'s representation is somewhat arduous, and their contribution is mostly focused on a rule-based workflow extraction framework rather than the assembly of a dataset that can be used by other methods. Neither \citeauthor{song_procedural_2011} nor \citeauthor{halioui_bioinformatic_2018}'s datasets are accessible.\footnote{\citeauthor{halioui_bioinformatic_2018} provide a link to their data and framework implementation in their paper, but the link is no longer active.}

\section{Dataset Collection Methodology and Overview}
Annotations for NER, NED, and other knowledge extraction tasks were curated by domain experts in computational biology for a series of 55 biomedical full-text papers and 1,195 abstracts, indexed on PubMed Central (PMC) and PubMed, respectively. We chose to include both full text and abstracts in FlaMB\'e to have a breadth of unique tokens as well as the depth needed to extract meaningful biological workflows.
  
\subsection{Collection Methodology}

\paragraph{Abstract corpus} The abstract corpus was hand-curated for tissue and cell type terms across 20 high-impact biomedical journals (full list in Supplementary Materials). To ensure that no single journal was overrepresented due to publication quantity, we set the number of sampled abstracts per journal to 60. Furthermore, we only sampled from recently published works between 2016 and 2021, as advances in technology have made it possible to study cell types in addition to bulk tissue and we want to capture the new diversity of cell types in our annotations. All abstracts were downloaded using PubMed eutils. To enable evaluation of interannotator agreement (Supplementary Materials), each of 3 annotators was assigned 400 abstracts (60 from each unique journal), with 240 overlapping abstracts evenly distributed across journals. 

\paragraph{Full text corpus} Because of the focus on single cell research, we used Pubmed eutils to query PMC for 3 general article types (``Classical Article,'' ``Clinical Study,'' and ``Journal Article'') using the following key words (allowing dashes to be used as a connector as well): ``scRNAseq,'' ``single cell RNAseq,'' ``single cell RNA sequencing,'' ``single cell transcriptomics,'' ``single cell transcriptome.'' Full text articles were downloaded directly via the PMC FTP and parsed using Pubmed Parser \cite{achakulvisut_pubmed_2020}. Out of the 55 total full text articles annotated by 2 annotators, 10 papers were annotated by both to evaluate interannotator agreement (Supplementary Materials).

\subsection{Annotation Types}

Tissue, cell type, tool, and method were annotated using the Prodigy software tool developed by Explosion AI for easy tracking of token-level tags. Due to the more limited presence of tool and methods, ergo tool context and workflow in abstracts, these annotations were only completed in the full text corpus. Tissue and cell type were annotated in both the abstract and full text corpora. 

\paragraph{Tissue and cell type}
To determine what classifies as tissue or cell type label, we use the terms in the NCI Thesaurus,\footnote{https://ncithesaurus.nci.nih.gov/ncitbrowser/} a comprehensive biomedical ontology for describing human samples which has cross-references to many other biomedical ontologies, as a guide. We focus on annotating useful sample descriptors that capture what biological entity is being studied, and try to tag the most specific term possible (e.g., ``left ventricle'' vs. ``ventricle''). The full set of annotation rules given to each annotator can be found in the Supplementary Materials.

A tissue or cell type in the text may be more specific than a term in the ontology, or it may not match exactly or any of the given synonyms. In these cases, we manually disambiguated the tag back to its nearest term in the ontology. In all other cases we programmatically mapped exact matches and synonyms back to NCIT identifiers. Additionally, in some cases, to express the specificity found in the text, we used two terms from the ontology in the disambiguation (e.g., 
``adipose stem cell'' is mapped to two terms in NCIT ``adipose'' and ``stem cell``). 

\paragraph{Tool and methods}
Unlike tissues and cell types which have standardized ontologies, there is no concrete vocabulary to annotate tools and methods in biomedical research. We have done our best to define two concrete categories of methods, those where an important computational transformation of the data has taken place but can be done by more than one package, (e.g., K-means clustering or PCA), and those that reference a specific tool or package. We label each of these respective types as unspecified method (``UNS\_METHOD'') or tool (``TOOL''). Furthermore, we aimed to identify computational methods applied on data that are separate from sequencing technologies and their related protocols (e.g., those done on machines which physically handle a biological sample) and only annotate tools and methods starting from the initial processing off of sequencing machines. 

\paragraph{Tool context}
In addition to annotating tools, methods, tissue, and cell type terms in the full text we also provide a set of tool ``contexts,'' or the analysis task that they are used for to process or augment data. This is important, as a single tool may have multiple functions or reasons that it was applied (Fig. 3 shows an example paper where Seurat used in 4 different contexts). For the sake of exploring the single cell multiverse, we restricted the set of modes to important functions in processing a wide variety of sequencing data. A single mention of a tool in the text can have one or more modes assigned to it based on its surrounding context. The full vocabulary for modes can be found in Supplementary Materials.

\paragraph{Workflow}
On a paper level, we aim to extract the various workflows done to samples, where samples are defined as an assay (e.g., scRNA-seq, ChIP-seq, BS-seq, etc) and a sample descriptor such as tissue/cell type pair. Once a unique set of samples per paper are identified we link them with tool and mode pairs from the text. Next, we annotate the flow by tabulating all edge pairs, where a pair of tools with their corresponding modes are applied to a given sample. In cases where an unspecified important transformation took place, such as an `UNS\_METHOD` we use ``unspecified\_mode'' as a placeholder. In this way we can reconstruct and model multiple workflows in a paper when more than one sample type is used.

\subsection{Dataset description and statistics}
\paragraph{Token level tags}
All token level tags, such as those for tissue and cell type and tool and method annotations are released as IOB and CoNLL files. The CoNLL files contain disambiguated annotations, with the tissue and cell type tags mapped semi-manually back to NCI Thesaurus identifiers and tools disambiguated back to a standardized name. An additional description file is also provided, one for tissues and cell types, which maps NCI Thesaurus ids to names, and one for tool and method annotations, with annotations to relevant references, GitHub, or project links.

Together, the full-text tag files span 55 papers and 429,373 tokens with 405 disambiguated (776 before disambiguation) tissue and cell type terms, 217 disambiguated tools, and 48 unique general methods. The abstract only tag files span 1,195 papers with 272,771 tokens annotated and 286 disambiguated tissue and cell type terms.

\paragraph{Tool context annotations}
Mode annotations for the various tools are provided in the tool and method CoNLL files. Each mode is manually assigned using the surrounding sentence context. 

\paragraph{Workflow annotations}
There is no predefined standard format for paper-level knowledge extraction annotations, so we split them into the following 3 files for easy parsing: A sample description and identification file, containing a listing of unique sample assay and tissue and cell type pairs; a tools applied file linking samples with the tool-mode combinations covering modes; and tool sequence file that ties pairs of tool-mode combinations together with sample identifiers. These files cover 8 unique assays, across 28 tool modes, capturing 390 tool-tool steps. There are on average, 10 workflow steps for each of the 38 papers with a defined workflow.

\section{FlaMB\'e Use Cases}
The diverse collection of annotations in FlaMB\'e enables several different use cases. We explore 3 example use cases of NER, tool context prediction, and workflow visualization before discussing other potential downstream applications.

\paragraph{Use case 1: named entity recognition}
We illustrate how the IOB and CoNLL files can be used to train BERT models to predict tissue and cell type mentions in biomedical abstracts. Using the full-text data as training and our abstract annotations as the hold out set for evaluation, we fine-tuned some of the most popular BERT models on HuggingFace (Table 1) for NER prediction. All models perform reasonably well, with PubMedBERT \cite{gu_domain-specific_2021} having the best F1 for the cell type and tissue type identification tasks. In general, the domain-specific pretrained language models do tend to perform better than the general domain models, especially when it comes to recall. 

\begin{table}
  \caption{\textbf{Predictive performance (P/R/F1 scores) of various language models on abstract tissue/cell type annotations.} Language models were fine-tuned on a combination of full text and abstracts and evaluated on a mixture of both text types for cell type and tissue annotations. Best performers are highlighted in bold.}
  \label{tab:tissue-benchmarking-f1}
  \centering
  \begin{tabular}{lcccccc}
      \toprule
      & \multicolumn{3}{c}{Cell Type} & \multicolumn{3}{c}{Tissue} \\ 
      \cmidrule(lr){2-4} 
      \cmidrule(lr){5-7}
    & Precision & Recall & F1 & Precision & Recall & F1 \\
     \midrule
    BERT-base \cite{devlin_bert_2019} & 0.740  & 0.823 & 0.779 & 0.813 & 0.811 & 0.812 \\
    ELECTRA \cite{clark_electra_2020}  & 0.775 & 0.802 & 0.788 & 0.819  & 0.841 & 0.830  \\
    BioBERT \cite{lee_biobert_2020}   & 0.725 & 0.806 & 0.763 & \textbf{0.848} & 0.859 & 0.854  \\
    BlueBERT \cite{peng_empirical_2020}    & 0.699 & 0.838 & 0.762 & 0.818 & 0.851 & 0.834 \\
    BioELECTRA \cite{kanakarajan_bioelectrapretrained_2021} & 0.719 & \textbf{0.855} & 0.781 & 0.751 & \textbf{0.894} & 0.816 \\
    PubMedBERT \cite{gu_domain-specific_2021} & \textbf{0.795} & 0.832 & \textbf{0.813} & 0.844 & 0.868 & \textbf{0.856}\\
    \bottomrule
  \end{tabular}
\end{table}

We also aim to demonstrate the utility of our annotations by comparing them with the only other easily obtainable software annotation dataset, Softcite \cite{du_softcite_2021}, a resource that provides annotations of software mentions in full-text research publications in the life sciences and economics. Here, we partition FlaMB\'e's full-text tool annotations into two sets of full-text data, holding out 11 randomly chosen papers for evaluation. We use the remaining 44 papers from FlaMB\'e and the entirety of Softcite for training. Both datasets were used to train PubMedBERT, one of the consistent performers in tissue/cell type prediction (Table 2). Despite being a smaller set of annotations, FlaMB\'e outperforms Softcite, especially when it comes to identifying the full name of a tool, (e.g., ``Search Tool for the Retrieval of Interacting Genes/Proteins,'' more commonly known as ``STRING''). This observation seems to be supported when we examine the predictive performance broken down by tag type---the largest performance difference between a model trained on Softcite and FlaMB\'e is in the `I-Tool' token (see Supplement). We hypothesize that the fact that biomedical tools often have long, multi-word names (and corresponding acronym) may play in role in this large difference. Of course, we note that in this comparison FlaMB\'e has the advantage of using the same annotation criteria in both the training and test sets, but nevertheless, we believe it still illustrates the importance and utility of FlaMB\'e's biomedical specific tool annotations.

\begin{table}
  \caption{\textbf{Predictive performance (F1 scores) of PubMedBERT on tool annotations when using Softcite or FlaMB\'e (excluding papers used for evaluation) as training standard.} Tool annotations from 11 full text papers were held out from FlaMB\'e as an evaluation standard. PubMedBERT was fine-tuned on either the entirety of Softcite annotations or the smaller FlaMB\'e training standard.}
  \label{tab:tool-dataset}
  \centering
  \begin{tabular}{lcccc}
    \toprule
       & Precision & Recall & F1\\
    \midrule
    Softcite \cite{du_softcite_2021} & 0.397 & 0.528 & 0.453   \\
    FlaMB\'e  & \textbf{0.779} & \textbf{0.909} & \textbf{0.839} \\
    \bottomrule
  \end{tabular}
\end{table}

\paragraph{Use case 2: tool context prediction}
As a proof of concept, we also used FlaMB\'e's tool context annotations and trained a PubMedBERT model to predict a tool's context given the sentence in which it is mentioned, akin to sentiment classification. We assembled a small set of training (191 sentences over 28 papers) and test (45 sentences over 8 papers) data, limiting ourselves to sentences containing a mention of at least one of the top 5 most mentioned tools, \textit{Seurat, Cellranger, t-SNE, Monocle}, and \textit{STAR}, each of which can be applied in multiple contexts. We then trained PubMedBERT models to predict context for each sentences in a one vs rest framework, for contexts that are well represented in the test and training datasets: \textit{Alignment, Marker Genes, and Clustering}. Each of the classifers performed well, with the alignment (AUC = 0.954) and marker gene (AUC = 0.953) contexts being more distinguishable and clustering (AUC = 0.810) being the most difficult. Given this promising performance on a test case, we anticipate that more sophisticated methods will be able to achieve consistently strong performance with our annotations.

\paragraph{Use case 3: visualization and exploration of different scientific workflows}
Different workflows can be extracted from FlaMB\'e's annotations, at different levels of specificity, either by highlighting the different tools used in a paper (Fig. 2A) or the different tool contexts in a paper (Fig. 2B). These can also be combined to extract more exact methodology (Fig. 3). Benchmarking papers or work introducing a new tool have to compare with previous work and create interesting workflows, as a small set of sample types is processed with slight variations through different levels of an entire pipeline depending on a paper's objective (Fig. 2). Meanwhile, papers that seek to solve a biological problem often have a more defined flow, with fewer tools from sample to one or more endpoints (Fig. 3). By extracting these workflows, we can not only classify the type of paper (e.g., benchmarking, new method, or biological insight), and analyze them on an individual level, but can also look at the global set of workflows for a large set of papers (Fig. 1). Thus, FlaMB\'e has important downstream potential for extracting knowledge at multiple levels.

\begin{figure}
  \centering
  \includegraphics[width=0.95\textwidth]{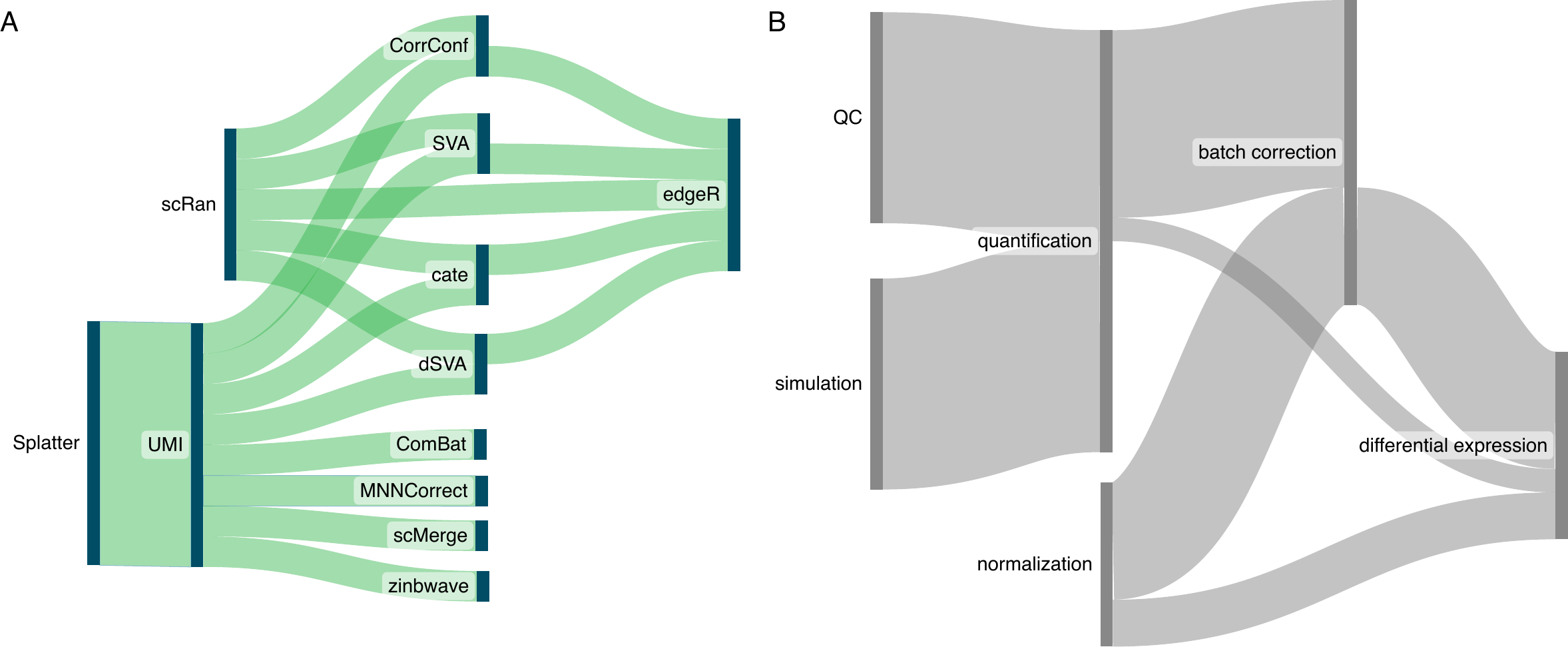}
  \caption{\textbf{Sankey visualizations of (A) tool and (B) context workflows from an example paper \cite{chen_comparison_2020}.} Visualizations here focus on one entity at a time, either the computational tools being used throughout the paper (vertical bars in A) or the context (vertical bars in B) in which they are being used.}
\end{figure}

\begin{figure}
  \centering
  \includegraphics[width=0.65\textwidth]{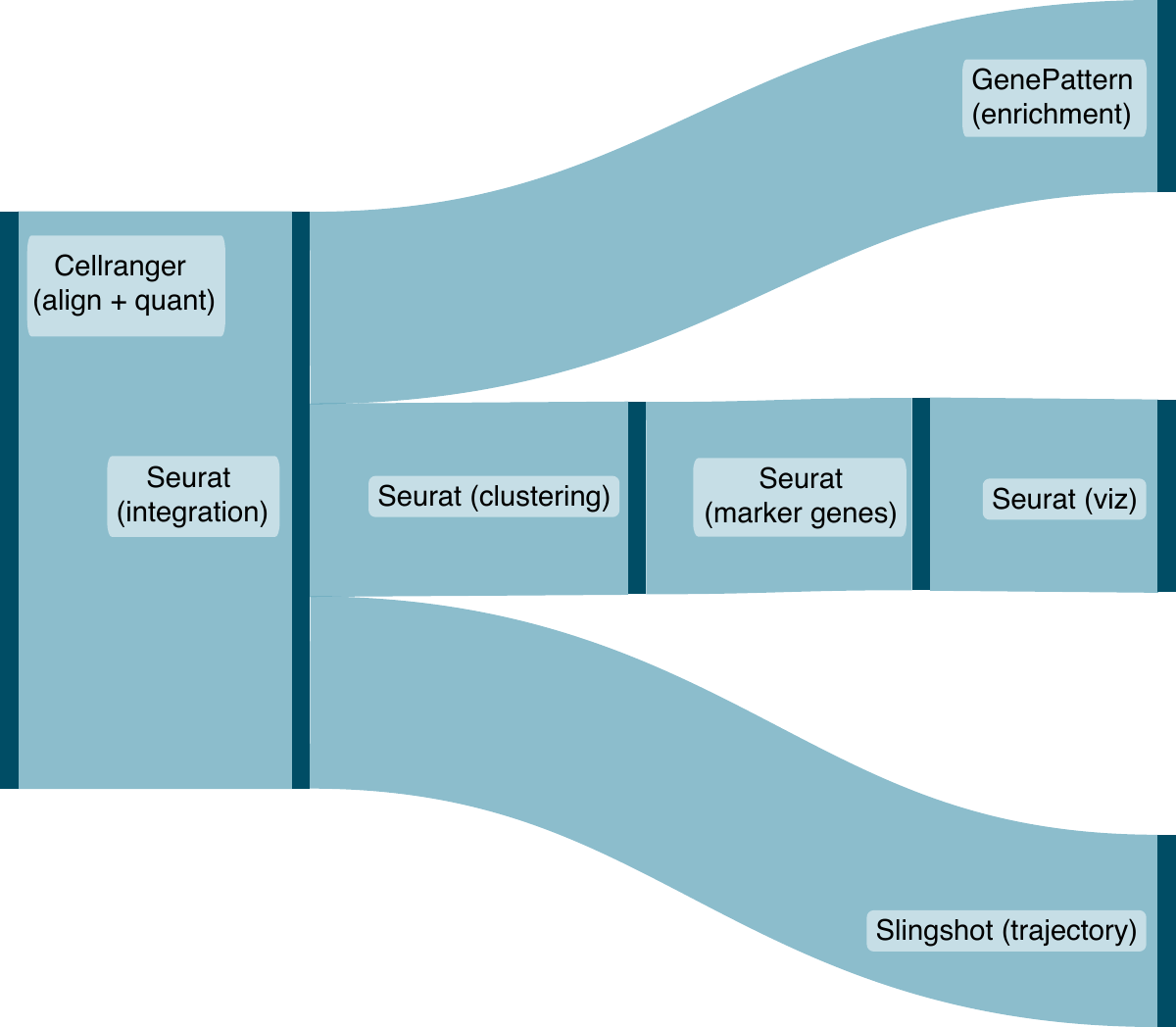 }
  \caption{\textbf{Sankey visualizations of the joint tool-context workflow from an example paper \cite{martos_single-cell_2020}.} Visualization here depicts the workflow of (tool, context) pairs (vertical bars), where context is denoted within the parentheses.}
\end{figure}

\subsection{Potential downstream applications}
There are many other interesting downstream applications that FlaMB\'e can be used to study. In addition to the advances in developing systematic methods for procedural knowledge extraction, we want to highlight the scientific value of improved modeling here. Specifically, structured representations would potentially allow for improved computational method recommendation depending on the goals of a particular study, as well as highlight gaps and areas of need for new computational method development. Importantly, one of the natural concerns that has been raised in psychology and more recently in machine learning is that having ever more complex computational workflows can spawn \textit{multiverses}. The multiverse represents the set of parallel universes where slightly different paths (e.g., methods or analysis steps) are taken towards the same goal. Multiverse analyses are undertaken to see how reliable results and conclusions are in light of these implicit decisions \cite{bell_modeling_2022,steegen_increasing_2016}. We believe one of the most exciting downstream applications of FlaMB\'e is systematic multiverse analyses of the complex workflows undertaken in biomedical research, towards the ultimate goal of improving transparency and reproducibility of research claims.

\section{Limitations and Future Work}
\label{limitations}
One of the current limitations of FlaMB\'e is that though the number of entity-level annotations is high, there are relatively fewer examples of the more complex annotation types of tool context and workflow. We plan to address this through larger annotation efforts that will further expand these categories. Because FlaMB\'e has also proposed a systematic, structured representation that can be used as input to existing language models, these future efforts can be aided by computational predictions that can guide manual curation efforts. In these follow-up efforts, we foresee that the NER-related annotations will be easiest to automate, followed be NED, with the tool context and workflow predictions being more challenging. Any automated annotations will be reviewed by expert curators before release of an updated dataset. We do not foresee negative societal impacts, though incorrect workflows could potentially be misleading for downstream research, and thus we would encourage thorough evaluation of all predictions.

With FlaMB\'e, we have broken down the more complex, abstract procedural knowledge extraction problem into more structured declarative knowledge tasks that the community is already well-equipped to tackle. Intriguingly, cognitive psychology research has pointed towards the fact that in humans, procedural and declarative knowledge are intertwined, but can sometimes be learned independently of one another \cite{willingham_development_1989}. Thus, there may also be benefit to using different, more ``procedural'' representations for learning. In some sense, one ML area that has tried to learn and mimic human procedural knowledge is reinforcement learning. A good example of this is with ``script knowledge'' \cite{schank_scripts_1975} and generally text-based games \cite{yuan_interactive_nodate}, which have used a game approach to improve modeling at the intersection of language understanding and complex decision-making. Reinforcement learning has also found some early success in reasoning over large scale knowledge graphs. Procedural knowledge extraction from academic texts could potentially also benefit from this type of framework. One of the unique aspects of FlaMB\'e is that though we have developed a structured representation, they can also tie together (e.g., we have annotated individual edges that can be viewed jointly as a graph). The disambiguated terms also tie in with existing knowledge bases that can be incorporated into knowledge graph research. It will be interesting to see whether new methods can be developed that could take advantage of the joint representation and learn more than the sum of the parts.

\section{Conclusion}
In conclusion, we have developed FlaMB\'e, a collection of datasets that together form structured representations of procedural knowledge captured in scientific literature. The dataset provides annotations for 1,195 paper abstracts and 55 full text papers, spanning nearly 700,000 tokens. In addition to providing the largest NER and NED dataset for tissue and cell type, we also provide annotations for computational tool and method, as well as the analysis task a tool is used in. Finally, we also annotate computational workflows within papers that can potentially be used in many downstream applications. Our dataset and associated code are accessible at \url{https://github.com/ylaboratory/flambe}.

\bibliographystyle{unsrtnat}
\bibliography{neurips_data_2023/references.bib}

\begin{thebibliography}{47}
\providecommand{\natexlab}[1]{#1}
\providecommand{\url}[1]{\texttt{#1}}
\expandafter\ifx\csname urlstyle\endcsname\relax
  \providecommand{\doi}[1]{doi: #1}\else
  \providecommand{\doi}{doi: \begingroup \urlstyle{rm}\Url}\fi

\bibitem[Nadeau and Sekine(2007)]{nadeau_survey_2007}
David Nadeau and Satoshi Sekine.
\newblock A survey of named entity recognition and classification.
\newblock \emph{Lingvisticæ Investigationes}, 30\penalty0 (1):\penalty0 3--26,
  January 2007.
\newblock ISSN 0378-4169, 1569-9927.
\newblock \doi{10.1075/li.30.1.03nad}.
\newblock URL
  \url{https://www.jbe-platform.com/content/journals/10.1075/li.30.1.03nad}.
\newblock Publisher: John Benjamins.

\bibitem[Li et~al.(2022)Li, Sun, Han, and Li]{li_survey_2022}
Jing Li, Aixin Sun, Jianglei Han, and Chenliang Li.
\newblock A {Survey} on {Deep} {Learning} for {Named} {Entity} {Recognition}.
\newblock \emph{IEEE Transactions on Knowledge and Data Engineering},
  34\penalty0 (1):\penalty0 50--70, January 2022.
\newblock ISSN 1558-2191.
\newblock \doi{10.1109/TKDE.2020.2981314}.
\newblock Conference Name: IEEE Transactions on Knowledge and Data Engineering.

\bibitem[Cucerzan(2007)]{cucerzan_large-scale_2007}
Silviu Cucerzan.
\newblock Large-{Scale} {Named} {Entity} {Disambiguation} {Based} on
  {Wikipedia} {Data}.
\newblock In \emph{Proceedings of the 2007 {Joint} {Conference} on {Empirical}
  {Methods} in {Natural} {Language} {Processing} and {Computational} {Natural}
  {Language} {Learning} ({EMNLP}-{CoNLL})}, pages 708--716, Prague, Czech
  Republic, June 2007. Association for Computational Linguistics.
\newblock URL \url{https://aclanthology.org/D07-1074}.

\bibitem[Zhang et~al.(2018)Zhang, Wang, and Liu]{zhang_deep_2018}
Lei Zhang, Shuai Wang, and Bing Liu.
\newblock Deep learning for sentiment analysis: {A} survey.
\newblock \emph{WIREs Data Mining and Knowledge Discovery}, 8\penalty0
  (4):\penalty0 e1253, 2018.
\newblock ISSN 1942-4795.
\newblock \doi{10.1002/widm.1253}.
\newblock URL \url{https://onlinelibrary.wiley.com/doi/abs/10.1002/widm.1253}.
\newblock \_eprint: https://onlinelibrary.wiley.com/doi/pdf/10.1002/widm.1253.

\bibitem[Zhang et~al.(2017)Zhang, Zhang, and Fu]{zhang_end--end_2017}
Meishan Zhang, Yue Zhang, and Guohong Fu.
\newblock End-to-{End} {Neural} {Relation} {Extraction} with {Global}
  {Optimization}.
\newblock In \emph{Proceedings of the 2017 {Conference} on {Empirical}
  {Methods} in {Natural} {Language} {Processing}}, pages 1730--1740,
  Copenhagen, Denmark, September 2017. Association for Computational
  Linguistics.
\newblock \doi{10.18653/v1/D17-1182}.
\newblock URL \url{https://aclanthology.org/D17-1182}.

\bibitem[Ryle(1945)]{ryle_knowing_1945}
Gilbert Ryle.
\newblock Knowing {How} and {Knowing} {That}: {The} {Presidential} {Address}.
\newblock \emph{Proceedings of the Aristotelian Society}, 46:\penalty0 1--16,
  1945.
\newblock ISSN 0066-7374.
\newblock URL \url{https://www.jstor.org/stable/4544405}.
\newblock Publisher: [Aristotelian Society, Wiley].

\bibitem[Wisdom(1949)]{wisdom_concept_1949}
John Wisdom.
\newblock The {Concept} of {Mind}.
\newblock \emph{Proceedings of the Aristotelian Society}, 50:\penalty0
  189--204, 1949.
\newblock ISSN 0066-7374.
\newblock URL \url{https://www.jstor.org/stable/4544471}.
\newblock Publisher: [Aristotelian Society, Wiley].

\bibitem[Georgeff and Lansky(1986)]{georgeff_procedural_1986}
M.P. Georgeff and A.L. Lansky.
\newblock Procedural knowledge.
\newblock \emph{Proceedings of the IEEE}, 74\penalty0 (10):\penalty0
  1383--1398, October 1986.
\newblock ISSN 1558-2256.
\newblock \doi{10.1109/PROC.1986.13639}.
\newblock Conference Name: Proceedings of the IEEE.

\bibitem[Mujtaba and Mahapatra(2019)]{mujtaba_recent_2019}
Dena Mujtaba and Nihar Mahapatra.
\newblock Recent {Trends} in {Natural} {Language} {Understanding} for
  {Procedural} {Knowledge}.
\newblock In \emph{2019 {International} {Conference} on {Computational}
  {Science} and {Computational} {Intelligence} ({CSCI})}, pages 420--424,
  December 2019.
\newblock \doi{10.1109/CSCI49370.2019.00082}.

\bibitem[Min et~al.(2019)Min, Jiang, Liu, Rui, and Jain]{min_survey_2019}
Weiqing Min, Shuqiang Jiang, Linhu Liu, Yong Rui, and Ramesh Jain.
\newblock A {Survey} on {Food} {Computing}.
\newblock \emph{ACM Computing Surveys}, 52\penalty0 (5):\penalty0 92:1--92:36,
  September 2019.
\newblock ISSN 0360-0300.
\newblock \doi{10.1145/3329168}.
\newblock URL \url{https://dl.acm.org/doi/10.1145/3329168}.

\bibitem[Chu et~al.(2017)Chu, Tandon, and Weikum]{chu_distilling_2017}
Cuong~Xuan Chu, Niket Tandon, and Gerhard Weikum.
\newblock Distilling {Task} {Knowledge} from {How}-{To} {Communities}.
\newblock In \emph{Proceedings of the 26th {International} {Conference} on
  {World} {Wide} {Web}}, {WWW} '17, pages 805--814, Republic and Canton of
  Geneva, CHE, April 2017. International World Wide Web Conferences Steering
  Committee.
\newblock ISBN 978-1-4503-4913-0.
\newblock \doi{10.1145/3038912.3052715}.
\newblock URL \url{https://dl.acm.org/doi/10.1145/3038912.3052715}.

\bibitem[Bellan et~al.(2023)Bellan, van~der Aa, Dragoni, Ghidini, and
  Ponzetto]{bellan_pet_2023}
Patrizio Bellan, Han van~der Aa, Mauro Dragoni, Chiara Ghidini, and
  Simone~Paolo Ponzetto.
\newblock {PET}: {An} {Annotated} {Dataset} for {Process} {Extraction}
  from {Natural} {Language} {Text} {Tasks}.
\newblock In Cristina Cabanillas, Niels~Frederik Garmann-Johnsen, and Agnes
  Koschmider, editors, \emph{Business {Process} {Management} {Workshops}},
  Lecture {Notes} in {Business} {Information} {Processing}, pages 315--321,
  Cham, 2023. Springer International Publishing.
\newblock ISBN 978-3-031-25383-6.
\newblock \doi{10.1007/978-3-031-25383-6_23}.

\bibitem[Gupta et~al.(2018)Gupta, Khosla, Singh, and
  Dasgupta]{gupta_mining_2018}
Abhirut Gupta, Abhay Khosla, Gautam Singh, and Gargi Dasgupta.
\newblock Mining {Procedures} from {Technical} {Support} {Documents}, May 2018.
\newblock URL \url{http://arxiv.org/abs/1805.09780}.
\newblock arXiv:1805.09780 [cs].

\bibitem[Methods(2014)]{nature_methods_method_2014}
Nature Methods.
\newblock Method of the {Year} 2013.
\newblock \emph{Nature Methods}, 11\penalty0 (1):\penalty0 1--1, January 2014.
\newblock ISSN 1548-7105.
\newblock \doi{10.1038/nmeth.2801}.
\newblock URL \url{https://www.nature.com/articles/nmeth.2801}.
\newblock Number: 1 Publisher: Nature Publishing Group.

\bibitem[Eberwine et~al.(2014)Eberwine, Sul, Bartfai, and
  Kim]{eberwine_promise_2014}
James Eberwine, Jai-Yoon Sul, Tamas Bartfai, and Junhyong Kim.
\newblock The promise of single-cell sequencing.
\newblock \emph{Nature Methods}, 11\penalty0 (1):\penalty0 25--27, January
  2014.
\newblock ISSN 1548-7105.
\newblock \doi{10.1038/nmeth.2769}.
\newblock URL \url{https://www.nature.com/articles/nmeth.2769}.
\newblock Number: 1 Publisher: Nature Publishing Group.

\bibitem[Tang et~al.(2009)Tang, Barbacioru, Wang, Nordman, Lee, Xu, Wang,
  Bodeau, Tuch, Siddiqui, Lao, and Surani]{tang_mrna-seq_2009}
Fuchou Tang, Catalin Barbacioru, Yangzhou Wang, Ellen Nordman, Clarence Lee,
  Nanlan Xu, Xiaohui Wang, John Bodeau, Brian~B. Tuch, Asim Siddiqui, Kaiqin
  Lao, and M.~Azim Surani.
\newblock {mRNA}-{Seq} whole-transcriptome analysis of a single cell.
\newblock \emph{Nature Methods}, 6\penalty0 (5):\penalty0 377--382, May 2009.
\newblock ISSN 1548-7105.
\newblock \doi{10.1038/nmeth.1315}.

\bibitem[Zappia et~al.(2018)Zappia, Phipson, and
  Oshlack]{zappia_exploring_2018}
Luke Zappia, Belinda Phipson, and Alicia Oshlack.
\newblock Exploring the single-cell {RNA}-seq analysis landscape with the
  {scRNA}-tools database.
\newblock \emph{PLOS Computational Biology}, 14\penalty0 (6):\penalty0
  e1006245, June 2018.
\newblock ISSN 1553-7358.
\newblock \doi{10.1371/journal.pcbi.1006245}.
\newblock URL
  \url{https://journals.plos.org/ploscompbiol/article?id=10.1371/journal.pcbi.1006245}.
\newblock Publisher: Public Library of Science.

\bibitem[Satija et~al.(2015)Satija, Farrell, Gennert, Schier, and
  Regev]{satija_spatial_2015}
Rahul Satija, Jeffrey~A. Farrell, David Gennert, Alexander~F. Schier, and Aviv
  Regev.
\newblock Spatial reconstruction of single-cell gene expression data.
\newblock \emph{Nature Biotechnology}, 33\penalty0 (5):\penalty0 495--502, May
  2015.
\newblock ISSN 1546-1696.
\newblock \doi{10.1038/nbt.3192}.
\newblock URL \url{https://www.nature.com/articles/nbt.3192}.
\newblock Number: 5 Publisher: Nature Publishing Group.

\bibitem[Beltagy et~al.(2019)Beltagy, Lo, and Cohan]{beltagy_scibert_2019}
Iz~Beltagy, Kyle Lo, and Arman Cohan.
\newblock {SciBERT}: {A} {Pretrained} {Language} {Model} for {Scientific}
  {Text}.
\newblock In \emph{Proceedings of the 2019 {Conference} on {Empirical}
  {Methods} in {Natural} {Language} {Processing} and the 9th {International}
  {Joint} {Conference} on {Natural} {Language} {Processing}
  ({EMNLP}-{IJCNLP})}, pages 3615--3620, Hong Kong, China, November 2019.
  Association for Computational Linguistics.
\newblock \doi{10.18653/v1/D19-1371}.
\newblock URL \url{https://aclanthology.org/D19-1371}.

\bibitem[Gu et~al.(2021)Gu, Tinn, Cheng, Lucas, Usuyama, Liu, Naumann, Gao, and
  Poon]{gu_domain-specific_2021}
Yu~Gu, Robert Tinn, Hao Cheng, Michael Lucas, Naoto Usuyama, Xiaodong Liu,
  Tristan Naumann, Jianfeng Gao, and Hoifung Poon.
\newblock Domain-{Specific} {Language} {Model} {Pretraining} for {Biomedical}
  {Natural} {Language} {Processing}.
\newblock \emph{ACM Transactions on Computing for Healthcare}, 3\penalty0
  (1):\penalty0 2:1--2:23, October 2021.
\newblock ISSN 2691-1957.
\newblock \doi{10.1145/3458754}.
\newblock URL \url{https://dl.acm.org/doi/10.1145/3458754}.

\bibitem[Lee et~al.(2020)Lee, Yoon, Kim, Kim, Kim, So, and
  Kang]{lee_biobert_2020}
Jinhyuk Lee, Wonjin Yoon, Sungdong Kim, Donghyeon Kim, Sunkyu Kim, Chan~Ho So,
  and Jaewoo Kang.
\newblock {BioBERT}: a pre-trained biomedical language representation model for
  biomedical text mining.
\newblock \emph{Bioinformatics}, 36\penalty0 (4):\penalty0 1234--1240, February
  2020.
\newblock ISSN 1367-4803.
\newblock \doi{10.1093/bioinformatics/btz682}.
\newblock URL \url{https://doi.org/10.1093/bioinformatics/btz682}.

\bibitem[Peng et~al.(2019)Peng, Yan, and Lu]{peng_transfer_2019}
Yifan Peng, Shankai Yan, and Zhiyong Lu.
\newblock Transfer {Learning} in {Biomedical} {Natural} {Language}
  {Processing}: {An} {Evaluation} of {BERT} and {ELMo} on {Ten} {Benchmarking}
  {Datasets}.
\newblock In \emph{Proceedings of the 18th {BioNLP} {Workshop} and {Shared}
  {Task}}, pages 58--65, Florence, Italy, August 2019. Association for
  Computational Linguistics.
\newblock \doi{10.18653/v1/W19-5006}.
\newblock URL \url{https://aclanthology.org/W19-5006}.

\bibitem[Kanakarajan et~al.(2021)Kanakarajan, Kundumani, and
  Sankarasubbu]{kanakarajan_bioelectrapretrained_2021}
Kamal~raj Kanakarajan, Bhuvana Kundumani, and Malaikannan Sankarasubbu.
\newblock {BioELECTRA}:{Pretrained} {Biomedical} text {Encoder} using
  {Discriminators}.
\newblock In \emph{Proceedings of the 20th {Workshop} on {Biomedical}
  {Language} {Processing}}, pages 143--154, Online, June 2021. Association for
  Computational Linguistics.
\newblock \doi{10.18653/v1/2021.bionlp-1.16}.
\newblock URL \url{https://aclanthology.org/2021.bionlp-1.16}.

\bibitem[Peng et~al.(2020)Peng, Chen, and Lu]{peng_empirical_2020}
Yifan Peng, Qingyu Chen, and Zhiyong Lu.
\newblock An {Empirical} {Study} of {Multi}-{Task} {Learning} on {BERT} for
  {Biomedical} {Text} {Mining}, May 2020.
\newblock URL \url{http://arxiv.org/abs/2005.02799}.
\newblock arXiv:2005.02799 [cs].

\bibitem[Weber et~al.(2021)Weber, Sänger, Münchmeyer, Habibi, Leser, and
  Akbik]{weber_hunflair_2021}
Leon Weber, Mario Sänger, Jannes Münchmeyer, Maryam Habibi, Ulf Leser, and
  Alan Akbik.
\newblock {HunFlair}: an easy-to-use tool for state-of-the-art biomedical named
  entity recognition.
\newblock \emph{Bioinformatics}, 37\penalty0 (17):\penalty0 2792--2794,
  September 2021.
\newblock ISSN 1367-4803.
\newblock \doi{10.1093/bioinformatics/btab042}.
\newblock URL \url{https://doi.org/10.1093/bioinformatics/btab042}.

\bibitem[Fries et~al.(2022)Fries, Weber, Seelam, Altay, Datta, Garda, Kang, Su,
  Kusa, Cahyawijaya, Barth, Ott, Samwald, Bach, Biderman, Sänger, Wang,
  Callahan, Periñán, Gigant, Haller, Chim, Posada, Giorgi, Sivaraman,
  Pàmies, Nezhurina, Martin, Cullan, Freidank, Dahlberg, Mishra, Bose, Broad,
  Labrak, Deshmukh, Kiblawi, Singh, Vu, Neeraj, Golde, Moral, and
  Beilharz]{fries_bigbio_2022}
Jason~Alan Fries, Leon Weber, Natasha Seelam, Gabriel Altay, Debajyoti Datta,
  Samuele Garda, Myungsun Kang, Ruisi Su, Wojciech Kusa, Samuel Cahyawijaya,
  Fabio Barth, Simon Ott, Matthias Samwald, Stephen Bach, Stella Biderman,
  Mario Sänger, Bo~Wang, Alison Callahan, Daniel~León Periñán, Théo
  Gigant, Patrick Haller, Jenny Chim, Jose~David Posada, John~Michael Giorgi,
  Karthik~Rangasai Sivaraman, Marc Pàmies, Marianna Nezhurina, Robert Martin,
  Michael Cullan, Moritz Freidank, Nathan Dahlberg, Shubhanshu Mishra, Shamik
  Bose, Nicholas~Michio Broad, Yanis Labrak, Shlok~S. Deshmukh, Sid Kiblawi,
  Ayush Singh, Minh~Chien Vu, Trishala Neeraj, Jonas Golde, Albert
  Villanova~del Moral, and Benjamin Beilharz.
\newblock {BigBio}: {A} {Framework} for {Data}-{Centric} {Biomedical} {Natural}
  {Language} {Processing}.
\newblock September 2022.
\newblock URL \url{https://openreview.net/forum?id=8lQDn9zTQlW}.

\bibitem[Ohta et~al.(2012)Ohta, Pyysalo, Tsujii, and
  Ananiadou]{ohta_open-domain_2012}
Tomoko Ohta, Sampo Pyysalo, Jun'ichi Tsujii, and Sophia Ananiadou.
\newblock Open-domain {Anatomical} {Entity} {Mention} {Detection}.
\newblock In \emph{Proceedings of the {Workshop} on {Detecting} {Structure} in
  {Scholarly} {Discourse}}, pages 27--36, Jeju Island, Korea, July 2012.
  Association for Computational Linguistics.
\newblock URL \url{https://aclanthology.org/W12-4304}.

\bibitem[Pyysalo and Ananiadou(2014)]{pyysalo_anatomical_2014}
Sampo Pyysalo and Sophia Ananiadou.
\newblock Anatomical entity mention recognition at literature scale.
\newblock \emph{Bioinformatics (Oxford, England)}, 30\penalty0 (6):\penalty0
  868--875, March 2014.
\newblock ISSN 1367-4811.
\newblock \doi{10.1093/bioinformatics/btt580}.

\bibitem[Pyysalo et~al.(2013)Pyysalo, Ohta, and
  Ananiadou]{pyysalo_overview_2013}
Sampo Pyysalo, Tomoko Ohta, and Sophia Ananiadou.
\newblock Overview of the {Cancer} {Genetics} ({CG}) task of {BioNLP} {Shared}
  {Task} 2013.
\newblock In \emph{Proceedings of the {BioNLP} {Shared} {Task} 2013
  {Workshop}}, pages 58--66, Sofia, Bulgaria, August 2013. Association for
  Computational Linguistics.
\newblock URL \url{https://aclanthology.org/W13-2008}.

\bibitem[Neves et~al.(2012)Neves, Damaschun, Kurtz, and
  Leser]{neves_annotating_2012}
Mariana Neves, Alexander Damaschun, Andreas Kurtz, and Ulf Leser.
\newblock Annotating and evaluating text for stem cell research.
\newblock January 2012.

\bibitem[Duck et~al.(2013)Duck, Nenadic, Brass, Robertson, and
  Stevens]{duck_bionerds_2013}
Geraint Duck, Goran Nenadic, Andy Brass, David~L. Robertson, and Robert
  Stevens.
\newblock {bioNerDS}: exploring bioinformatics’ database and software use
  through literature mining.
\newblock \emph{BMC Bioinformatics}, 14\penalty0 (1):\penalty0 194, June 2013.
\newblock ISSN 1471-2105.
\newblock \doi{10.1186/1471-2105-14-194}.
\newblock URL \url{https://doi.org/10.1186/1471-2105-14-194}.

\bibitem[Du et~al.(2021)Du, Cohoon, Lopez, and Howison]{du_softcite_2021}
Caifan Du, Johanna Cohoon, Patrice Lopez, and James Howison.
\newblock Softcite dataset: {A} dataset of software mentions in biomedical and
  economic research publications.
\newblock \emph{Journal of the Association for Information Science and
  Technology}, 72\penalty0 (7):\penalty0 870--884, 2021.
\newblock ISSN 2330-1643.
\newblock \doi{10.1002/asi.24454}.
\newblock URL \url{https://onlinelibrary.wiley.com/doi/abs/10.1002/asi.24454}.
\newblock \_eprint: https://onlinelibrary.wiley.com/doi/pdf/10.1002/asi.24454.

\bibitem[Schindler et~al.(2021)Schindler, Bensmann, Dietze, and
  Krüger]{schindler_somesci-_2021}
David Schindler, Felix Bensmann, Stefan Dietze, and Frank Krüger.
\newblock {SoMeSci}- {A} 5 {Star} {Open} {Data} {Gold} {Standard} {Knowledge}
  {Graph} of {Software} {Mentions} in {Scientific} {Articles}, August 2021.
\newblock URL \url{http://arxiv.org/abs/2108.09070}.
\newblock arXiv:2108.09070 [cs].

\bibitem[Istrate et~al.(2022)Istrate, Li, Taraborelli, Torkar, Veytsman, and
  Williams]{istrate_large_2022}
Ana-Maria Istrate, Donghui Li, Dario Taraborelli, Michaela Torkar, Boris
  Veytsman, and Ivana Williams.
\newblock A large dataset of software mentions in the biomedical literature,
  September 2022.
\newblock URL \url{http://arxiv.org/abs/2209.00693}.
\newblock arXiv:2209.00693 [cs, q-bio].

\bibitem[Schindler et~al.(2022)Schindler, Bensmann, Dietze, and
  Krüger]{schindler_role_2022}
David Schindler, Felix Bensmann, Stefan Dietze, and Frank Krüger.
\newblock The role of software in science: a knowledge graph-based analysis of
  software mentions in {PubMed} {Central}.
\newblock \emph{PeerJ Computer Science}, 8:\penalty0 e835, January 2022.
\newblock ISSN 2376-5992.
\newblock \doi{10.7717/peerj-cs.835}.
\newblock URL \url{https://peerj.com/articles/cs-835}.
\newblock Publisher: PeerJ Inc.

\bibitem[Song et~al.(2011)Song, Oh, Myaeng, Choi, Chun, Choi, and
  Jeong]{song_procedural_2011}
Sa-kwang Song, Heung-seon Oh, Sung~Hyon Myaeng, Sung-pil Choi, Hong-woo Chun,
  Yun-soo Choi, and Chang-hoo Jeong.
\newblock Procedural {Knowledge} {Extraction} on {MEDLINE} {Abstracts}.
\newblock In Ning Zhong, Vic Callaghan, Ali~A. Ghorbani, and Bin Hu, editors,
  \emph{Active {Media} {Technology}}, Lecture {Notes} in {Computer} {Science},
  pages 345--354, Berlin, Heidelberg, 2011. Springer.
\newblock ISBN 978-3-642-23620-4.
\newblock \doi{10.1007/978-3-642-23620-4_36}.

\bibitem[Halioui et~al.(2018)Halioui, Valtchev, and
  Diallo]{halioui_bioinformatic_2018}
Ahmed Halioui, Petko Valtchev, and Abdoulaye~Baniré Diallo.
\newblock Bioinformatic {Workflow} {Extraction} from {Scientific} {Texts} based
  on {Word} {Sense} {Disambiguation}.
\newblock \emph{IEEE/ACM Transactions on Computational Biology and
  Bioinformatics}, 15\penalty0 (6):\penalty0 1979--1990, November 2018.
\newblock ISSN 1557-9964.
\newblock \doi{10.1109/TCBB.2018.2847336}.
\newblock Conference Name: IEEE/ACM Transactions on Computational Biology and
  Bioinformatics.

\bibitem[Achakulvisut et~al.(2020)Achakulvisut, Acuna, and
  Kording]{achakulvisut_pubmed_2020}
Titipat Achakulvisut, Daniel~E. Acuna, and Konrad Kording.
\newblock Pubmed {Parser}: {A} {Python} {Parser} for {PubMed} {Open}-{Access}
  {XML} {Subset} and {MEDLINE} {XML} {Dataset} {XML} {Dataset}.
\newblock \emph{Journal of Open Source Software}, 5\penalty0 (46):\penalty0
  1979, February 2020.
\newblock ISSN 2475-9066.
\newblock \doi{10.21105/joss.01979}.
\newblock URL \url{https://joss.theoj.org/papers/10.21105/joss.01979}.

\bibitem[Devlin et~al.(2019)Devlin, Chang, Lee, and
  Toutanova]{devlin_bert_2019}
Jacob Devlin, Ming-Wei Chang, Kenton Lee, and Kristina Toutanova.
\newblock {BERT}: {Pre}-training of {Deep} {Bidirectional} {Transformers} for
  {Language} {Understanding}, May 2019.
\newblock URL \url{http://arxiv.org/abs/1810.04805}.
\newblock arXiv:1810.04805 [cs].

\bibitem[Clark et~al.(2020)Clark, Luong, Le, and Manning]{clark_electra_2020}
Kevin Clark, Minh-Thang Luong, Quoc~V. Le, and Christopher~D. Manning.
\newblock {ELECTRA}: {Pre}-training {Text} {Encoders} as {Discriminators}
  {Rather} {Than} {Generators}, March 2020.
\newblock URL \url{http://arxiv.org/abs/2003.10555}.
\newblock arXiv:2003.10555 [cs].

\bibitem[Chen et~al.(2020)Chen, Zhang, Williams, Ju, Shaner, Easton, Wu, and
  Chen]{chen_comparison_2020}
Wenan Chen, Silu Zhang, Justin Williams, Bensheng Ju, Bridget Shaner, John
  Easton, Gang Wu, and Xiang Chen.
\newblock A comparison of methods accounting for batch effects in differential
  expression analysis of {UMI} count based single cell {RNA} sequencing.
\newblock \emph{Computational and Structural Biotechnology Journal},
  18:\penalty0 861--873, 2020.
\newblock ISSN 2001-0370.
\newblock \doi{10.1016/j.csbj.2020.03.026}.

\bibitem[Martos et~al.(2020)Martos, Campbell, Lozoya, Wang, Bennett, Thompson,
  Wan, Pittman, and Bell]{martos_single-cell_2020}
Suzanne~N. Martos, Michelle~R. Campbell, Oswaldo~A. Lozoya, Xuting Wang,
  Brian~D. Bennett, Isabel J.~B. Thompson, Ma~Wan, Gary~S. Pittman, and
  Douglas~A. Bell.
\newblock Single-cell analyses identify dysfunctional {CD16}+ {CD8} {T} cells
  in smokers.
\newblock \emph{Cell Reports. Medicine}, 1\penalty0 (4):\penalty0 100054, July
  2020.
\newblock ISSN 2666-3791.
\newblock \doi{10.1016/j.xcrm.2020.100054}.

\bibitem[Bell et~al.(2022)Bell, Kampman, Dodge, and
  Lawrence]{bell_modeling_2022}
Samuel~J. Bell, Onno Kampman, Jesse Dodge, and Neil Lawrence.
\newblock Modeling the {Machine} {Learning} {Multiverse}.
\newblock \emph{Advances in Neural Information Processing Systems},
  35:\penalty0 18416--18429, December 2022.
\newblock URL
  \url{https://proceedings.neurips.cc/paper_files/paper/2022/hash/750337e1301941f81ae31a90e0a1c181-Abstract-Conference.html}.

\bibitem[Steegen et~al.(2016)Steegen, Tuerlinckx, Gelman, and
  Vanpaemel]{steegen_increasing_2016}
Sara Steegen, Francis Tuerlinckx, Andrew Gelman, and Wolf Vanpaemel.
\newblock Increasing transparency through a multiverse analysis.
\newblock \emph{Perspectives on Psychological Science}, 11:\penalty0 702--712,
  2016.
\newblock ISSN 1745-6924.
\newblock \doi{10.1177/1745691616658637}.
\newblock Place: US Publisher: Sage Publications.

\bibitem[Willingham et~al.(1989)Willingham, Nissen, and
  Bullemer]{willingham_development_1989}
Daniel~B. Willingham, Mary~J. Nissen, and Peter Bullemer.
\newblock On the development of procedural knowledge.
\newblock \emph{Journal of Experimental Psychology: Learning, Memory, and
  Cognition}, 15:\penalty0 1047--1060, 1989.
\newblock ISSN 1939-1285.
\newblock \doi{10.1037/0278-7393.15.6.1047}.
\newblock Place: US Publisher: American Psychological Association.

\bibitem[Schank and Abelson(1975)]{schank_scripts_1975}
Roger~C. Schank and Robert~P. Abelson.
\newblock Scripts, plans, and knowledge.
\newblock In \emph{Proceedings of the 4th international joint conference on
  {Artificial} intelligence - {Volume} 1}, {IJCAI}'75, pages 151--157, San
  Francisco, CA, USA, September 1975. Morgan Kaufmann Publishers Inc.

\bibitem[Yuan()]{yuan_interactive_nodate}
Matthew Hausknecht{\textbar}{\textbar}Prithviraj
  Ammanabrolu{\textbar}{\textbar}Marc-Alexandre~Côté{\textbar}{\textbar}Xingdi
  Yuan.
\newblock Interactive {Fiction} {Games}: {A} {Colossal} {Adventure}.
\newblock URL
  \url{https://aaai.org/papers/07903-interactive-fiction-games-a-colossal-adventure/}.

\end{thebibliography}

\end{document}